%% file: egpaper_for_review.tex
\documentclass[10pt,twocolumn,letterpaper]{article}

\usepackage{iccv}
\usepackage{times}
\usepackage{epsfig}
\usepackage{amsmath}
\usepackage{amssymb}

\usepackage{amsmath}
\usepackage{amssymb}
\usepackage{booktabs}

\usepackage{algorithm}
\usepackage{multicol}
\usepackage{algpseudocode}
\usepackage{caption}

\usepackage{amsthm}

\usepackage[utf8]{inputenc} 
\usepackage[T1]{fontenc}    
\usepackage{url}            
\usepackage{booktabs}       
\usepackage{amsfonts}       
\usepackage{nicefrac}       
\usepackage{microtype}      
\usepackage{xcolor}         
\usepackage{microtype}
\usepackage{graphicx}
\usepackage{booktabs} 
\usepackage{multirow}
\usepackage{comment}
\usepackage{xspace}
\newcommand{\wrong}[1]{\textcolor{red}{}}

\newcommand{\ourmethod}{\textbf{AdvMacer \xspace}}
\newcommand{\EsbRs}[1]{\textbf{EsbRs#1}}
\DeclareMathOperator*{\argmax}{\arg\max}

\newcommand\mb{\mathbb}
\newcommand\mE{\mathbb{E}}
\newcommand\mP{\mathbb{P}}

\newtheorem{theorem}{Theorem}[section]

\newtheorem{remark}[theorem]{Remark}

\usepackage[pagebackref=true,breaklinks=true,letterpaper=true,colorlinks,bookmarks=false]{hyperref}
\usepackage[capitalize,noabbrev]{cleveref}
\crefname{section}{Sec.}{Secs.}
\Crefname{section}{Section}{Sections}
\Crefname{table}{Table}{Tables}
\crefname{table}{Tab.}{Tabs.}

\iccvfinalcopy 

\def\iccvPaperID{9627} 

\ificcvfinal\fi

\begin{document}

\title{Promoting Robustness of Randomized Smoothing: \\ Two Cost-Effective Approaches}

\author{Linbo Liu\thanks{Work done prior to joining Amazon. This manuscript is the extended version of the ICDM 23 paper.}\\
AWS AI Labs, UC San Diego\\
{\tt\small linbol@amazon.com}
\and
Trong Nghia Hoang\\
Washington State University\\
{\tt\small trongnghia.hoang@wsu.edu}
\and
Lam M. Nguyen\\
IBM Research\\
{\tt\small LamNguyen.MLTD@ibm.com}
\and
Tsui-Wei Weng\\
UC San Diego\\
{\tt\small lweng@ucsd.edu}
}

\maketitle
\ificcvfinal\thispagestyle{empty}\fi

\begin{abstract}
Randomized smoothing has recently attracted attentions in the field of adversarial robustness to provide provable robustness guarantees on smoothed neural network classifiers. However, existing works show that \textit{vanilla} randomized smoothing usually does not provide good robustness performance and often requires (re)training techniques on the base classifier in order to boost the robustness of the resulting smoothed classifier. In this work, we propose two cost-effective approaches to boost the robustness of randomized smoothing while preserving its clean performance. The first approach introduces a new robust training method \ourmethod which combines adversarial training and robustness certification maximization for randomized smoothing. We show that \ourmethod can improve the robustness performance of randomized smoothing classifiers compared to SOTA baselines, while being 3$\times$ faster to train than MACER baseline. The second approach introduces a post-processing method \textbf{EsbRS} which greatly improves the robustness certificate based on building model ensembles. We explore different aspects of model ensembles that has not been studied by prior works and propose a novel design methodology to further improve robustness of the ensemble based on our theoretical analysis. 
\end{abstract}

\input{1_intro}
\input{2_related_work}
\input{3_ours_method}

\input{4_simu}
\input{5_conclusion}

{\small
\bibliographystyle{ieee_fullname}
\bibliography{egbib}
}

\newpage
\appendix
\input{appendix}

\end{document}

%% file: 1_intro.tex
\section{Introduction}

The existence of adversarial examples of deep neural networks (DNNs)~\cite{szegedy2014intriguing,goodfellow2015explaining} has raised serious concerns to deploy DNNs in real-world systems, especially in the safety critical applications such as self-driving cars and aircraft control systems. Thus, many research efforts have been devoted into developing effective defenses methods to safeguard DNNs. One of the most promising direction is known as \textit{certified defense} via \textit{randomized smoothing}, where the word \textit{certified} means that the defense methods have provable theoretical guarantee as opposed to easily broken heuristic defenses~\cite{athalye2018obfuscated}, and \textit{randomized smoothing} is a popular technique that allows scalable certified defenses for state-of-the-art DNNs against adversarial examples. Randomized smoothing is recently proposed by \cite{Lecuyer2019Certified,li2018attacking,cohen2019certified} and has achieved state-of-the-art robustness guarantees. Given any classifier $f$, denoted as a \textit{base classifier}, randomized smoothing predicts the most-likely class on the randomly perturbed input $x$ with Gaussian noises. Following this new prediction rule, randomized smoothing acts as an operator on the original \textit{base classifier} and produce a new \textit{smoothed} classifier which comes with provable robustness guarantees under various $\ell_p$ norm threat models \cite{Li2019Certified,cohen2019certified}.


\begin{figure}
    \centering
    \includegraphics[width=0.8\columnwidth]{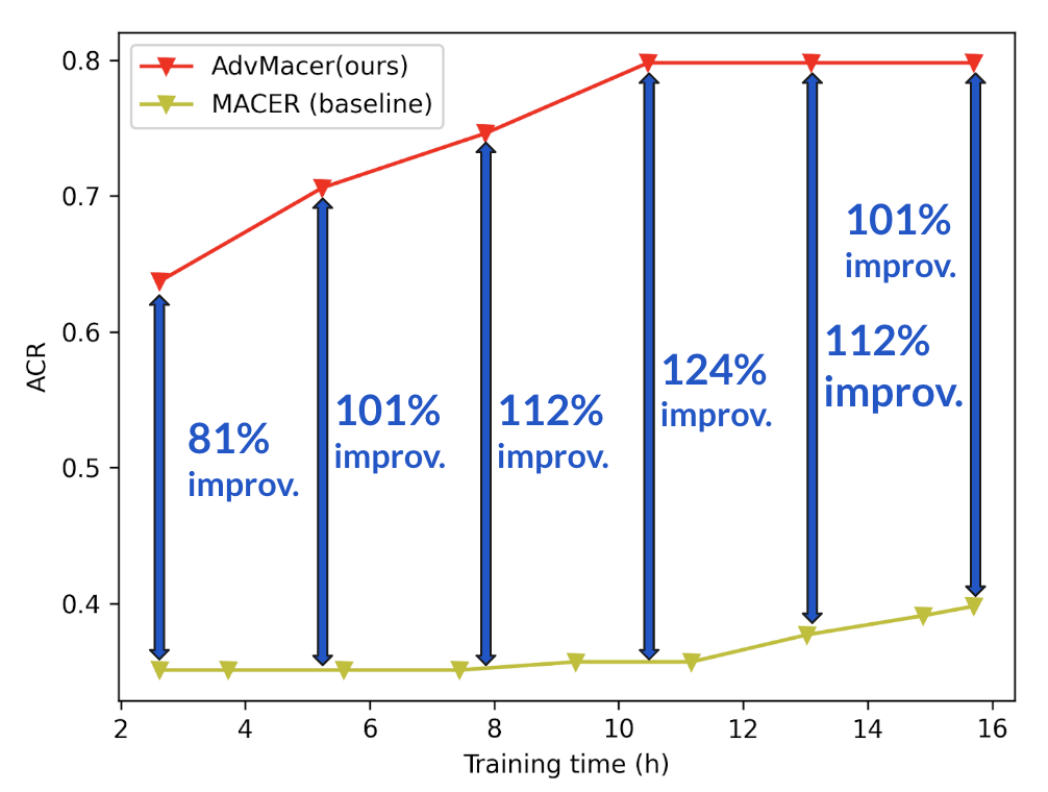}
    \caption{Best ACR seen so far: $x$-axis is the training time in hours and $y$-axis is the best ACR obtained before this specific training time. ACR is recorded for every 25 training epochs. Our proposed method is plotted in red, while the baseline is in light yellow. The experiments are on Cifar-10 with $\sigma=1.00$. Improvement over baseline model is indicated in blue.}
    \label{fig:acr-plot}
\end{figure}

Unfortunately, without specially-designed training techniques, the robustness certificate of the smoothed classifier is usually very weak~\cite{cohen2019certified}. Thus, a few recent works~\cite{salman2019provably,zhai2019macer} have proposed specialized robust training algorithms to improve robustness of the smoothed classifier. In~\cite{salman2019provably}, the authors propose an adversarial training method called SmoothAdv, which is similar to the PGD training~\cite{madry2018towards} but on the \textit{smoothed} classifier. On the other hand, \cite{zhai2019macer} propose MACER, whose training objective involves a term to maximize the robustness certificate directly. However, SmoothAdv often requires heavy tuning on a number of hyper-parameters for different noise level $\sigma$, which could be computationally challenging; while MACER usually requires longer (3$\times$) training epochs to train and unfortunately the resulting models often have weaker certificate despite higher clean accuracy. 

Motivated by the need of cost-effective robust training methods for randomized smoothing, we propose two approaches to address the limitations of existing robust training algorithms. Our contributions are three-fold: \textbf{1)} we propose a new robust training method called \ourmethod, which takes the best of both worlds in SmoothAdv and MACER: \ourmethod can achieve the best ACR while having the same computational cost as SmoothAdv and much (3$\times$) faster than MACER. If compared with MACER under same training time, our \ourmethod shows remarkable improvement (up to 124\%) in ACR over MACER, which is illustrated in \cref{fig:acr-plot}.
\textbf{2)} we equip our \ourmethod models with a training-free ensemble method \EsbRs{}, which can further enlarge the resulting model's certified radius (by up to 8\% compared with SmoothAdv and 15\% compared with MACER). Crucially, we present a general theoretical analysis on ensembles and demonstrate the effect of both \textit{intra}-model ensembles and \textit{mixed}-model ensembles from the theoretical point of view. \textbf{3)} grounded by our theoretical findings, an optimal weighted ensemble can be derived analytically where the weights are dependent on the input data. 

\wrong{Third, inspired by our heuristic finding, a novel 0/1-weighted ensemble named \EsbRs{-}max is proposed where the weights are data-dependent but noise-independent. We note that \ourmethod and \EsbRs{} are \textit{integrated} contributions as \ourmethod not only works well by itself, but can also help \EsbRs{} to boost up the robustness of ensemble model even more.}

\begin{figure*}[t!]
\begin{center}
\centerline{\includegraphics[width=400pt]{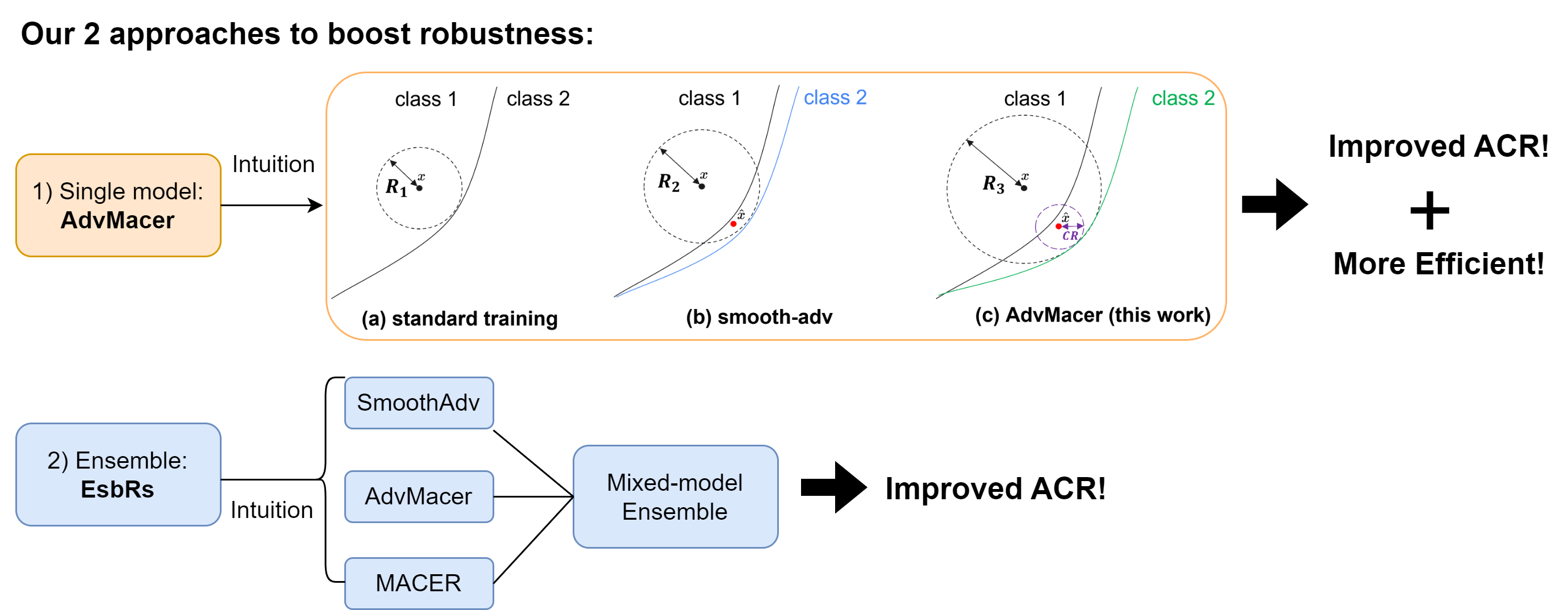}}
\caption{The overview figure, including the illustration of the idea behind \ourmethod: $x$ (black dot) is the original data sample and $\hat x$ (red dot) is an adversarial example of $x$. The solid black line is the original decision boundary. The blue line in (b) is the decision boundary using SmoothAdv and the green line in (c) is the decision boundary after applying \ourmethod. SmoothAdv force the classifier to classify $\hat x$ correctly to get the red boundary. \ourmethod force $\hat x$ to not only have correct prediction but also a large margin. Therefore, \ourmethod can obtain larger certified radius $R_3$ $>$ certified radius of smoothadv $R_2$ $>$ certified radius of the original classifier $R_1$.}
\label{fig:intuition}
\end{center}
\vskip -0.2in
\end{figure*}

%% file: 2_related_work.tex
\section{Related works and backgrounds}


In this section, we first give backgrounds on randomized smoothing and the related certified defense SmoothAdv~\cite{salman2019provably} and MACER~\cite{zhai2019macer}. Next, we review recent liteature on applying ensemble methods to randomized smoothing. 

\paragraph{Randomized smoothing.} Consider a neural network classifier $f:\mb{R}^d\to\mathcal{Y}$ that maps an input sample $x\in\mb{R}^d$ to its predicted label in $\mathcal{Y}$. \cite{cohen2019certified} introduced a randomized smoothing (RS) technique that can turn any base classifier $f(x)$ into a 
smoothed classifier $g(x)$ with provable robustness guarantees. When taking a sample $x$, the smoothed classifier $g$ returns the class that the base classifier $f$ is most likely to return under isotropic Gaussian noise perturbation of $x$:
$
    g(x) = \argmax_{c\in\mathcal{Y}}\mP_{\epsilon\sim\mathcal{N}(0,\sigma^2I)}(f(x+\epsilon)=c),
$
where $\sigma$ is the noise level that controls the trade-off between clean accuracy and model robustness. 
\cite{cohen2019certified} further proved the robustness guarantees of such smoothed classifier in Theorem~\ref{thm:cohen}. Let $\Phi$ denote the cumulative density function (CDF) of the standard Gaussian distribution. Suppose that under Gaussian perturbation $\epsilon\sim\mathcal{N}(0, \sigma^2I)$, the most likely class $c_A$ is returned with probability $p_A$ and the second most likely (runner-up) class $c_B$ is returned with probability $p_B$, i.e. 
$
    c_A=\argmax_{c\in\mathcal{Y}}\mP(f(x+\epsilon)=c),
    c_B=\argmax_{c\neq c_A}\mP(f(x+\epsilon)=c),
    p_A=\mP(f(x+\epsilon)=c_A), \, p_B=\mP(f(x+\epsilon)=c_B).
$
\begin{theorem}[Theorem 1 of \cite{cohen2019certified}]
\label{thm:cohen}
Assume $p_A$ attains a lower bound $\underbar p_A$ and $p_B$ attains an upper bound $\bar p_B$ with $\underbar p_A>\bar p_B$, then $g(x+\delta)=c_A$ for all $\|\delta\|_2 < R$, where 
$
        R = \frac\sigma2(\Phi^{-1}({\underbar p_A})-\Phi^{-1}({\bar p_B})).
$
\end{theorem}
In practice, Monte Carlo sampling is employed to obtain an estimate of $p_A$, see \cite{cohen2019certified}.
Unfortunately, as reported in ~\cite{cohen2019certified}, the robustness certificate is weak without any specifically-designed training techniques for randomized smoothing. To enhance the robustness of randomized smoothing, \cite{salman2019provably} proposed to train base classifier $f$ on adversarial examples of soft-RS classifiers $G(x)$, which are generated by PGD \cite{madry2018towards}. Another line of work \cite{zhai2019macer} considered an attack-free robust training by directly maximizing certified radius of each training sample. We briefly revisit these two methods \cite{salman2019provably, zhai2019macer} in the following: Formally, let $F^\beta:\mb{R}^d\to P(\mathcal{Y})$ be the soft version of classifier $f$ whose last layer is a softmax layer with inverse temperature $\beta$ and $P(\mathcal{Y})$ is a probability distribution over the label space $\mathcal{Y}$. We omit the superscript $\beta$ if there is no ambiguity and denote a smoothed soft classifier as
$G(x)=\mE_{\delta\sim\mathcal{N}(0,\sigma^2I)}F(x+\delta).$

\paragraph{SmoothAdv.} \cite{salman2019provably} introduced SmoothAdv to find adversarial examples by PGD. Denote $L_{\text{CE}}$ as the canonical cross entropy loss. Given a labeled data $(x, y)$, SmoothAdv finds a point $\hat x$ that maximizes the cross entropy loss of $G(x)$ in the local neighborhood of $x$:
\begin{align}
\label{eq:pgd}
    \hat x &=\argmax_{\|x'-x\|_2\leq\epsilon}L_{\text{CE}}(G(x'), y)\notag\\
    &=\argmax_{\|x'-x\|_2\leq\epsilon}-\log\mE_{\delta\sim\mathcal{N}(0, \sigma^2 I)} F(x'+\delta)_y.
\end{align}
Such optimization problem \eqref{eq:pgd} is solved by projected gradient descent (PGD). To estimate the gradient of \eqref{eq:pgd}, \cite{salman2019provably} used Monte Carlo simulation to approximate $\nabla_{x'}L_{\text{CE}}(G(x'), y)$ by 
$\nabla_{x'}\Big(-\log\Big(\frac1m\sum_{k=1}^mF(x'+\delta_k)_y\Big)\Big),$
where $\delta_1,\dots,\delta_m$ are drawn i.i.d. from $\mathcal{N}(0, \sigma^2I)$.

\paragraph{MACER.} Since the certified radius is related to the difference between the top probability $p_A$ and the runner-up probability $p_B$, \cite{zhai2019macer} constructed MACER loss $L_{\text{MACER}}$, which aims at simultaneously minimizing classification error and maximizing the certified radius of correctly classified samples. Specifically, 
\begin{align}
\label{eq:macer}
L_{\text{MACER}}(x)&=L_{\text{CE}}(G(x), y)+\lambda L_{\text{R}}(G;x, y),
\end{align}
where $\lambda\geq0$ is a tuning parameter. The loss in \eqref{eq:macer} involves the soft smoothed classifier $G$ and \cite{zhai2019macer} proposes to approximate $G(x)$ by Monte Carlo sampling:
\begin{align}
\label{eq:hatz}
    G(x)&\approx \hat z(x)=\frac1m\sum_{k=1}^mF(x+\delta_k),\notag\\
    \hat g(x)&=\argmax_{i\in\mathcal{Y}}\hat z_i(x).
\end{align}
where $\delta_1,\dots,\delta_m$ are drawn i.i.d. from $\mathcal{N}(0, \sigma^2I)$. Denote $\widehat{\text{CR}}(x,y)$ as the approximated certified radius at $x$, and 
$\widehat{\text{CR}}(x,y)=\frac\sigma2(\Phi^{-1}(\hat z_y(x)) - \Phi^{-1}(\max_{y'\neq y}\hat z_{y'}(x))).$
Therefore, the robustness loss $L_{\text{R}}(G;x,y)$ can also be approximated by
\begin{align}
\label{eq:LR}
    L_{\text{R}}(\hat z;x,y)
    &=\phi(\epsilon+\tilde\epsilon-\widehat{\text{CR}}(x, y))\,\textbf{1}(\hat g(x)=y)\notag\\
    &=\frac{\sigma}2 \phi(\gamma-\hat\xi_{\theta}(x, y))\,\textbf{1}(\hat g(x)=y),
\end{align}
where $\epsilon,\tilde\epsilon>0$ are hyper-parameters in the hinge loss, $\phi(u)=\max\{u,0\}$, $\gamma=\frac2{\sigma}(\epsilon+\tilde\epsilon)$,
and
$
    \hat\xi_\theta(x, y)=\Phi^{-1}(\hat z_y(x)) - \Phi^{-1}(\max_{y'\neq y}\hat z_{y'}(x)).
$
Finally, MACER trains a base classifier by minimizing the approximated MACER loss on training dataset. We refer readers to \cite{zhai2019macer} for more details. 

\paragraph{Other related works.} 
\cite{jeong2020consistency} introduced consistency loss as a regularization to improve the robustness of RS classifiers. Also, the usage of mixture of adversarial examples and clean examples is suggested in training to increase robust certification as in \cite{jeong2021smoothmix}. As can be shown in \cref{sec:simu}, our proposed \ourmethod outperforms all baselines on Cifar-10.

\paragraph{Boosting robustness via ensemble.} Model ensemble is a popular technique in the machine learning literature to practically improve model performance and reduce generalization errors~\cite{allen2020towards}. Recently, there are a few works investigating the idea of using model ensemble to improve robustness of a randomized smoothed classifier~\cite{horvath2021boosting, yang2021certified}. However, \cite{horvath2021boosting} mainly focused on ensemble the same type of models (i.e. models trained from the same process but with different random seeds) and only gave a brief exploration (in their App. G3.5) of mixed-model ensemble.
In contrast, as will be introduced in Section 3.2, our proposed \EsbRs{} is a more general ensemble method where we theoretically analyze the effect of \textit{mixed}-model ensembles. Although weighted ensemble has also been studied in \cite{horvath2021boosting}, their model learns the weights from training and cannot justify the weights' optimality. However, in our work, we develop a novel design framework of the optimal weight ensemble by solving an optimization problem based on our theory.
\wrong{
Although \cite{yang2021certified} also studied 0/1 weights in Max-Margin-Ensemble (MME), their weights are noise-\textit{dependent} while ours are noise-\textit{independent}. We note that this seemingly small difference actually imposes a big difference in both the robustness performance and computational cost (see \cref{subsubsec:esb-max} for detailed discussion). In fact, their MME design is not optimal in the sense that MME only achieves comparable performance as naive average ensemble and that extra cost is needed in computing max margin for each noisy samples. In contrast, our \EsbRs{-}max has significant improvement over average ensemble and even reduces the certification cost.}

%% file: 3_ours_method.tex
\section{Our proposed main methods}

In this section, we propose two novel and cost-effective approaches to improve robustness of a randomized smoothed classifier. First, we introduce a new robust training method \ourmethod that aim to maximize the certified radius over adversarial examples, while being 3$\times$ faster to train than MACER due to faster convergence. We present the intuitions, formulations as well as the details of our algorithm in \cref{subsec:maceradv}. Next, in \cref{subsec:ensemble}, we propose a novel ensemble method called \EsbRs{} with theoretical analysis.
Different from the two recent works~\cite{yang2021certified,horvath2021boosting}, we provide a more general analysis which does not require individual classifiers to come from the same training method. Our analysis allows the derivation of the optimal weight for individual classifiers, which is the key to promote robustness and the study of optimal weight has not been explored in the prior work. 

\wrong{Our analysis provides insights into how to design an optimal data-dependent but noise-independent 0/1-weighted ensemble, which has not been explored in the prior work.}

\vspace{-3mm}
\subsection{Approach 1: \ourmethod}
\label{subsec:maceradv}
Inspired by the prior work SmoothAdv~\cite{salman2019provably} and MACER~\cite{zhai2019macer} and to address their limitations, we argue that a smoothed classifier can be trained to have larger certified radius by directly optimizing the certified radius of adversarial examples instead of the clean data points. Notice that this statement requires adversarial example to be predicted correctly, which is encoded in our $L_{\text{CE}}$ term (without this constraint, the certified radius of original data point may actually decrease). The intuition is illustrated in \cref{fig:intuition}. Based on the above idea, we propose the following formulation. 
\vspace{-3mm}
\paragraph{Formulation.} Given data $x$ and its label $y$, we aim to minimize the proposed \ourmethod loss consisting of two terms:
$
    L_{\text{AdvMacer}}(x)=L_{\text{CE}}(\hat z(\hat x), y) + \lambda L_{\text{R}}(\hat z;\hat x, y),
$
where $\hat z$ and $L_{\text{R}}$ are given in \cref{eq:hatz} and \cref{eq:LR} respectively. The 1st term $L_{\text{CE}}(\hat z(\hat x), y)$ is to encourage adversarial examples $\hat x$ to be classified correctly, and the 2nd term 
$L_{\text{R}}(\hat z;\hat x, y)=\frac{\sigma}2 \max\{\gamma-\hat\xi_{\theta}(\hat x, y), 0\}\,\textbf{1}(\hat g(\hat x)=y)$
is to maximize the certified radius at the adversarial example $\hat x$, where 
$\hat x=\argmax_{\|x'-x\|_2\leq\epsilon}L_{\text{CE}}(\hat z(x'), y).$
To minimize the $L_{\text{AdvMacer}}(x)$, we generate the adversarial examples $\hat x$ via \cref{eq:pgd} with $T$-step PGD using SmoothAdv~\cite{salman2019provably}, i.e. in the $i$-th step, we update 
$
    x_{i+1} = \prod_{\mathcal{B}(x,\epsilon)}(x_i + \nabla_{x}(-\log(\frac1m\sum_{k=1}^mF(x+\delta_k)_y))|_{x=x_i}),
$
where $\prod_\mathcal{S}(\cdot)$ is the projection onto set $\mathcal{S}$ and we set $\hat x=x_T$. The training objective is to minimize $L_{\text{AdvMacer}}(x)$ by first-order optimization method, and a detailed algorithm is presented in \cref{app:alg} due to page constraint.





\vspace{-3mm}
\paragraph{Hyper-parameters.} Note that there are a few hyper-parameters in \ourmethod: $\sigma$ is the noise level that is introduced when $f$ or $F$ is smoothed; $\epsilon$ in \cref{eq:pgd} controls the size of the $\ell_2$ ball when doing PGD; $\gamma$ in \cref{eq:LR} is the parameter in hinge loss; $\lambda$ is the regularization parameter which controls the trade-off between clean accuracy and robustness; $m$ in \cref{eq:hatz} is the number of Monte Carlo samples used to estimate $G(x)$; $T$ is the number of PGD step to generate adversarial samples. Finally, recall that the soft classifier $F=F^{\beta}$, where $\beta$ is the inverse temperature in softmax layer. The larger $\beta$ is, the closer the soft classifier $F$ is to the hard classifier $f$.

\vspace{-3mm}
\paragraph{Discussion and Comparison.} \textcolor{blue}{\textbf{(I).}} Our \ourmethod is NOT just a naive combination of existing work. Instead, this clever observation has intuition (as shown in \cref{fig:intuition}) and can achieve better results (as shown in Table 1). 
Our proposed \ourmethod trains a model on adversarial examples while taking certified radius into consideration, which bridged between robust training and adversarial training. \textcolor{blue}{\textbf{(II). }}Compared with SmoothAdv, \ourmethod doesn't bring any additional computational overhead to calculate robust loss as there exist analytic formula for certified radius; in the meantime, compared with MACER, we require much fewer number of epochs (3$\times$ smaller) and faster training time (2-4 $\times$ faster) to obtain a robust model with much larger certified radius. From the experiments in \cref{sec:simu}, it can be seen that \ourmethod outperforms both SmoothAdv and MACER on various dataset, such as Cifar-10, ImageNet and SVHN. \textcolor{blue}{\textbf{(III). }}Equipped with our ensemble method \EsbRs{} presented in \cref{subsec:ensemble}, \ourmethod also enriches the diversity of component models, making mixed-model ensemble more robust (see \cref{fig:ensemble}). For a thorough comparison by experiments, see \cref{sec:simu} for more details.

\subsection{Approach 2: \EsbRs{}}
\label{subsec:ensemble}
\subsubsection{Analysis}
Ensemble is a cost-effective post-training technique to enhance model performance and reduce generalization error without spending much additional efforts on re-training the neural networks. By simply averaging the output from several models, ensemble shows remarkable boost in test accuracy and model robustness. Recently, there are a few works investigating the idea of using model ensemble to improve robustness of a randomized smoothed classifier~\cite{horvath2021boosting,yang2021certified}. However, the existing work mainly focused on ensembling similar classifiers (\textbf{intra-model ensemble}) with naive averaged weights. In contrast, we also consider \textbf{mixed-model ensemble} with component classifiers coming from different training methods and conduct theoretical analysis explaining the success of mixed ensemble in certain cases. Besides, unlike~\cite{liu2020enhancing} learning the ensemble weights empirically from training set, we develop a novel theoretical framework to design optimal ensemble weights based on our analysis. Empirical experiments verify the superiority of our proposed methods.
\wrong{Moreover, we introduce a 0/1-weight design strategy called \EsbRs{-}max to further boost the robustness of \EsbRs{} and reduce computational cost.}



\vspace{-3mm}
\paragraph{Formulation.}\quad Suppose we have $k$ trained soft classifiers $F^1,\dots,F^k:\mb{R}^d\to P(\mathcal{Y})$ and $\mathcal{Y}=\{1,\dots,c\}$. Consider soft-ensemble model $H$ whose output is a weighted average of the probabilities from $F^1,\dots,F^k$:
$H(x)=\sum_{l=1}^kw_l(x)F^l(x).$ Suppose the associated hard classifier is
$h(x)=\argmax_{c\in\mathcal{Y}}\big(H(x)\big)_c.$ Note that the weights here can be either data-dependent or data-independent. When it comes to data-dependent randomized smoothing, another line of work \cite{alfarra2022data,eiras2021ancer} consider the noise level $\sigma$ being data-dependent, which is different from this work. A more closely related work \cite{yang2021certified} proposed Max-Margin-Ensemble (MME) that has data-dependent weight function. A comparison of our method to MME is summarized in \cref{table_wt_ensemble}.
Then we apply RS to $h$ and get the corresponding smoothed classifier $e$. It can be shown that the RS classifier $e$ of an ensemble model also has the same certification guarantee as in \cite{cohen2019certified}, which is presented in the following theorem.

\begin{theorem}[Robustness gaurantee for ensemble]\label{thm_ours}
Suppose that under Gaussian perturbation $\epsilon\sim\mathcal{N}(0, \sigma^2I)$, the most likely class $c_A$ is returned by the 
\EsbRs{} $e$ with probability $p_A$ and the second most likely class $c_B$ is returned with probability $p_B$,
then we have $e(x+\delta)=e(x)$ for all $\|\delta\|_2\leq R$, where $R = \frac\sigma2(\Phi^{-1}({ p_A})-\Phi^{-1}({ p_B})).$
\end{theorem}
The proof is given in \cref{app_proof}.
Extensive experiments from \cref{sec:simu} show that ensemble-RS (\EsbRs{}) classifier $g$ noticeably improves both accuracy and robustness, no matter $F^l$ comes from the same or different training methods. Specifically, if $F^l$ comes from more than one training methods, we call $g$ a \textbf{mixed-model ensemble}.

\paragraph{Theoretical analysis.}\quad We present some theoretical analysis on how mixed-model ensemble can reduce the variance and hence increase certified radius. We generalize the analysis in \cite{horvath2021boosting} to allow mixed ensemble, which provide deeper insights on model ensemble study.

For a fixed query point $x$ with a Gaussian perturbation $\epsilon\sim\mathcal{N}(0,\sigma^2I)$, suppose probability logits vector $\boldsymbol y^l\in P(\mathcal{Y})$ is returned by $F^l$. Without loss of generality, assume 1 is the majority class in RS for the ensemble model $h(x)$. For simplicity, we can work with classification margin $z^l_i=y^l_1-y^l_i$, for $i\in\mathcal{Y}$. Let $\bar {\boldsymbol y}=H(x+\epsilon)$. Therefore, $\bar {\boldsymbol y}=\sum_{l=1}^kw_l\boldsymbol y^l.$ Similarly define $\bar z_i=\bar y_1-\bar y_i$. 
Consider $\mE[\bar {\boldsymbol z}]\in\mb{R}^c$ and 
$\text{Var}(\bar {\boldsymbol z})\in\mb{R}^{c\times c}$
, where the expectation is taken over the randomness in training process, including random initialization and stochasticity in GD. Then we have
\begin{align}
\label{eq:varz}
    \text{Var}(\bar {\boldsymbol z}) &= \text{Var}\Big(\sum_{l=1}^kw_l\boldsymbol z^l\Big)\notag\\
    &=\sum_{l=1}^kw^2_l\text{Var}(\boldsymbol z^l) + 2\sum_{l\neq m}w_lw_m\text{Cov}(\boldsymbol z^l, \boldsymbol z^m)
\end{align}
Hence, $\text{Var}(\bar z_i) = \text{Var}(\bar {\boldsymbol z})_{ii}$. Denote $p_i(w)=\text{Var}(\bar z_i)$ as a function of $w=[w_1,\dots,w_k]^\top$. Suppose there are a fixed number of training methods and denote this number by $s$, so  
$
    \alpha_i=\alpha_i(s)=\max_{1\leq l\leq k}\text{Var}(\boldsymbol z^l)_{ii},
    \beta_i=\beta_i(s)=\max_{l\neq m}\text{Cov}(\boldsymbol z^l, \boldsymbol z^m)_{ii}
$
are functions of $s$ instead.
As a result, $\alpha_i(s),\beta_i(s)=O(1)$ even as $k\to\infty$. 

\paragraph{A special case.}\quad As a special case, consider $w_l=\frac1k$ for all $l=1,\dots,k$. By \cref{eq:varz}, we derive 
\begin{equation}
\label{eq:varzi}
    p_i(w)=\text{Var}(\bar z_i)\leq\frac{k\alpha_i+k(k-1)\beta_i}{k^2}=\beta_i+\frac{\alpha_i-\beta_i}{k}.
\end{equation}
These classifiers either come from different training methods, or same training method with different random seeds. Thus, existing work all assumes 
that the logits from one classifier have larger covariance $\alpha_i$ than the logits from different classifiers $\beta_i$. However, as we will see in \textbf{Discussion} paragraph, ensemble may harm the performance if the above assumption doesn't hold. For now, let's assume $\alpha_i>\beta_i$. By \cref{eq:varzi}, we conclude that the upper bound of $\text{Var}(z_i)$ decreases to a constant $\beta_i$ as $k\to\infty$. 

Next, we explain how $\text{Var}(\bar z_i)$ affects certified radius. From \cref{thm:cohen}, we see that $R=\sigma\Phi^{-1}(\underbar p_A)$ if $\underbar p_A\geq\frac12$, hence we only need to show a lower bound on the top class probability $p_A$ increases as $k$ becomes larger. Since we assume the majority class's number is 1, we see that

    $p_1=\mP(\bar z_i>0, \forall i=2,\dots,c) \geq1-\sum_{i=2}^c\mP(\bar z_i\leq0).$

By Chebyshev's inequality, $\mP(\bar z_i\leq0)\leq \mP\Big(\big|\bar z_i-\mE[\bar z_i]\big|\geq\mE[\bar z_i]\Big)\leq \frac{\text{Var}(\bar z_i)}{\mE[\bar z_i]^2}$ and let $e_i=e_i(s)=\min_{l}\mE[z^l_i]$, thus we have
\begin{equation}\label{eq:lower_bound}
        p_1\geq1-\sum_{i=2}^c\frac{\text{Var}(\bar z_i)}{e_i^2}.
\end{equation}

The above equation suggests us to choose the weight $w$ that maximizes the RHS of \cref{eq:lower_bound} to have a larger $p_1$, hence larger certified radius. Since $e_i$ is independent of the choice of $w$,  we can obtain the optimal weight by solving
\begin{align}
\label{eq:opt_wt}
    \min_{w\in\mb{R}^k}\sum_{i=2}^ca_ip_i(w) \;\;\; \text{s.t.} \sum_{l=1}^kw_l = 1,\quad w_l\geq0,
\end{align}
where $a_i=e_i^{-2}$ are constants. Note that when $w_l=\frac1k$ for all $l=1,\dots, k$, we have a lower bound on $p_1$ by \eqref{eq:varzi} and \eqref{eq:lower_bound}:
\begin{align*}
    p_1\geq1-\sum_{i=2}^c\frac{\beta_i+(\alpha_i-\beta_i)/k}{e_i^2} \to1-\sum_{i=2}^c\frac{\beta_i}{e_i^2}\;\;\text{as }k\to\infty.
\end{align*}
This explains why larger $k$ makes $p_1$ and certified radius larger even in average ensemble.

\begin{figure}[t]
\vskip 0.2in
\begin{center}
\centerline{\includegraphics[width=200pt]{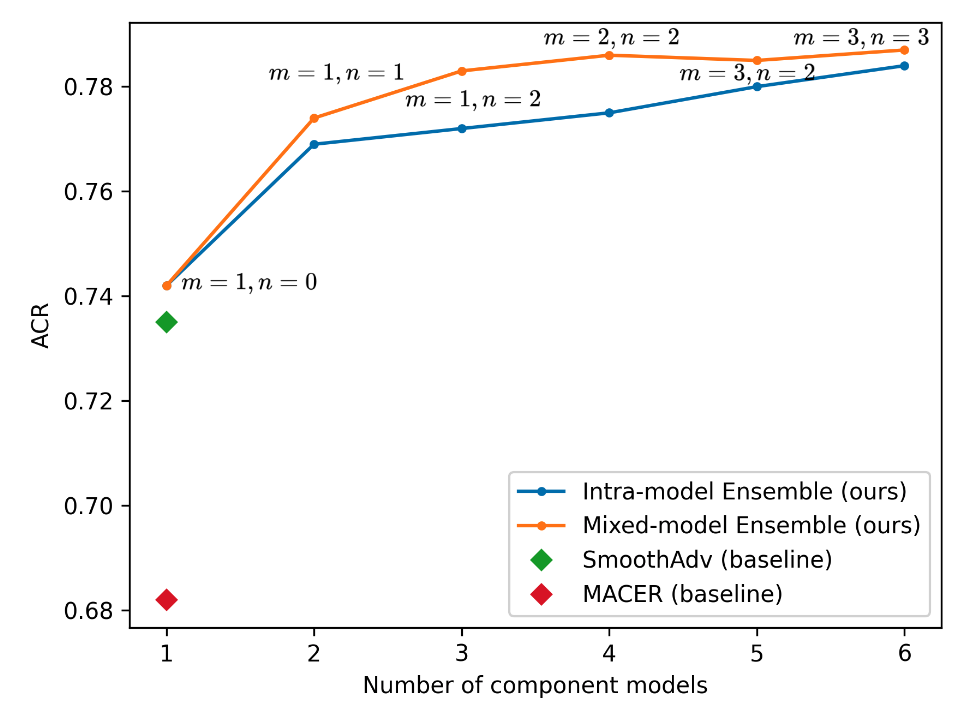}}
\caption{The plot of ACR against different number of component models in \EsbRs{} on Cifar-10 with $\sigma=0.50$. Single ensemble uses $N$ \ourmethod models. Mixed ensemble with totally $N$ component models uses $m$ \ourmethod models and $n$ SmoothAdv models.}
\label{fig:ensemble}
\end{center}
\vskip -0.2in
\end{figure}

\paragraph{Discussion.} Compared with \cite{horvath2021boosting}, we generalize their analysis to allow mixed-model ensemble and hence have several new findings. \textcolor{blue}{\textbf{(I). }} If $\alpha_i<\beta_i$, namely the logits from one model have smaller variance than those from different models, the RHS of \cref{eq:varzi} becomes an increasing function in $k$, which implies ensemble does not always work. \textcolor{blue}{\textbf{(II). }} We are the first to provide \textbf{both} theoretical analysis and comprehensive investigation on the advantage of a {mixed-model ensemble}: Suppose $F^1,F^2$ come from model category 1 (for example, SmoothAdv) and $F^3$ comes from model category 2 (for example, \ourmethod). If the logits from different types of models have smaller variance than those from the same type of model, namely $\text{Cov}(F^1,F^3)<\text{Cov}(F^1,F^2)$,
$\beta_i$ will become smaller and makes mixed-model ensemble work better than intra-model ensemble. This phenomenon is observed in \cref{fig:ensemble}. \cref{fig:ensemble} shows potential benefits that could be brought by heterogeneity of ensemble models, which is in accordance with our theoretical analysis presented in this section. 
\textcolor{blue}{\textbf{(III). }} \cite{horvath2021boosting} only gave a brief exploration (in their App. G3.5) suggesting similar ensemble would outperform mixed-model ensemble and did not further analyze the reasons. Our work pointed out that it's NOT always the case: we had in fact established concrete scenarios where \textbf{mixed-model} ensemble would have \textbf{benefit} over a intra-model ensemble both theoretically 
and empirically (see below \cref{fig:ensemble}). It serves as a counter example to the statement in Appendix G3.5 of \cite{horvath2021boosting}. Importantly, we note that our theoretical analysis is \textit{more general} than that presented in \cite{horvath2021boosting} and would reduce to theirs if setting $\text{cov}(z^l,z^m)=\text{var}(z^l)$ in \cref{eq:varz}. 

\paragraph{Designing optimal weighted ensemble.}
The optimization problem in \eqref{eq:opt_wt} allows us to design an optimal weight that can maximize the lower bound on $p_1$. Consider the case where $k=2$, then \eqref{eq:opt_wt} can be solved analytically given the knowledge of $\text{Var}(z^1), \text{Var}(z^2), \text{Cov}(z^1, z^2)$ and $a_i$. To see this, let $b_i=\text{Var}(z^1)_{ii}, c_i=2\text{Cov}(z^1,z^2)_{ii},d_i=\text{Var}(z^2)_{ii}$, then the objective function in \eqref{eq:opt_wt} can be re-written as 
\begin{align}
    q(w)&=\sum_{i=2}^ca_i(b_iw_1^2+c_iw_1w_2+d_iw_2^2)\notag\\
    &\stackrel{\text{(i)}}{=}\sum_{i=2}^c a_i[b_iw_1^2+c_iw_1(1-w_1)+d_i(1-w_1)^2],
\end{align}
where (i) uses the constraint $w_1+w_2=1$ to eliminate $w_2$. Therefore, the problem \eqref{eq:opt_wt} can be further cast as a quadratic optimization with linear constraints:
\begin{align}
\label{eq:q}
    \min_{w_1\in\mb{R}}\quad&q(w_1)=Aw_1^2+Bw_1+C\notag\\
    \text{s.t.}\quad& 0\leq w_1\leq1,
\end{align}
where $A=\sum_{i=2}^ca_i(b_i+c_i+d_i),B=\sum_{i=2}^c -a_i(c_i+2d_i)$ and $C=\sum_{i=2}^ca_id_i$. 
Notice that this problem has an analytical solution: if $A>0$ and $0\leq-\frac{B}{2A}\leq1$, $w_1=-\frac{B}{2A}$ and $w_2=1+\frac{B}{2A}$; else $q(w_1)$ attains minimum at boundary $w_1=0\text{ or }1$.

Next, we aim at giving an estimate of $a_i,b_i,c_i,d_i$. To account for randomness both from training and Gaussian perturbation $\epsilon$ around the input $x$, we first generate $n$ i.i.d. Gaussian noisy data $x_1,\dots,x_n$ from $\mathcal{N}(x,\sigma^2I)$. Second, we incorporate random perturbation for the parameters $\theta$ in classifier $F$ to imitate random seeds in training, as this is the cheapest way (without extra training cost). We randomly select $t\%$ parameters from $F$ and add i.i.d. Gaussian noise $\delta\sim\mathcal{N}(0,\tilde\sigma^2)$ for each selected parameter. This returns a perturbed model $\hat F$ from the base model $F$. Repeating the above process on $F^1$ and $F^2$ for $m$ times gives us $2m$ perturbed models $\hat F_1^1,\dots,\hat F^1_m$ and $\hat F^2_1,\dots,F^2_m$. 

Now, we pass $x_1,\dots,x_n$ into $\hat F^1_1,\dots,\hat F^1_m$ to get $mn$ output logits vector $y^{1,1},y^{1,2},\dots,y^{1,mn}$. Also, $y^{2,1},y^{2,2},\dots,y^{2,mn}$ can be obtained similarly by passing $n$ noisy data into $m$ perturbed models of $F^2$. Compute $z^{l,j}_i=y^{l,j}_1-y^{l,j}_i$ for $1\leq i\leq c$ and $l=1,2$.
Then an estimation of variance and covariance can be their empirical parallel:
\begin{align*}
    b_i&=\text{Var}(z^1)_{ii}=\frac1{mn}\sum_{j=1}^{mn}(z^{1,j}-\bar z^{1})(z^{1,j}-\bar z^{1})^\top_{ii},\\
    c_i&=2\text{Cov}(z^1,z^2)_{ii}=\frac2{mn}\sum_{j=1}^{mn}(z^{1,j}-\bar z^{1})(z^{2,j}-\bar z^{2})^\top_{ii},\\
    d_i&=\text{Var}(z^2)_{ii}=\frac1{mn}\sum_{j=1}^{mn}(z^{2,j}-\bar z^{2})(z^{2,j}-\bar z^{2})^\top_{ii},\\
\end{align*}
where $\bar z^l=\frac1{mn}\sum_{j=1}^{mn}z^{l,j}$ for $l=1,2$. Also obtain $a_i=e_i^{-2}=\min\{\bar z^1_i,\bar z^2_i\}^{-2}$
Hence, we can solve \eqref{eq:q} by plugging in $a_i,b_i,c_i,d_i$. A detailed algorithm is given in \cref{alg:wt_ensemble} in \cref{app:wt_ensemble}.

\begin{remark}
\textcolor{blue}{\textbf{(I).}} To our best knowledge, we are the first work to develop a practical and theoretical grounded methodology to obtain the optimal weight of the ensemble scheme by solving a optimization problem. We note that the two recent works~\cite{yang2021certified,horvath2021boosting} did not explore this direction. \textcolor{blue}{\textbf{(II).}} Our design strategy can be easily generalized to $k>2$. However, due to non-convexity of the objective function in \cref{eq:opt_wt}, solving the optimization problem \eqref{eq:opt_wt} is always intractable. Thus we only provide empirical experiments on $k=2$ and will leave $k>2$ as a future work.
\end{remark}

%% file: 4_simu.tex
\section{Experiments}
\label{sec:simu}

\input{tables/table_main_result_cifar}
In this section, we present experimental results that empirically evaluate the performance of our proposed methods, \ourmethod and \EsbRs{}, on Cifar-10 \cite{krizhevsky2009learning}, ImageNet \cite{deng2009imagenet} and SVHN \cite{netzer2011reading} dataset. To make fair comparisons with previous baseline models, we use the same architectures as in \cite{cohen2019certified}: ResNet-110 \cite{he2016deep}. We train our models with $\sigma=0.25,0.50,1.00$ on Cifar-10 and ImageNet, and $\sigma=0.25,0.50$ on SVHN. We train all models on a single NVIDIA V100 GPU and the training time reported below is all from NVIDIA V100 GPU. We compare the performance of \ourmethod with four baseline models (SmoothAdv, MACER, Consistency, SmoothMix) in \cref{table_main} and observe that our \ourmethod achieves the largest ACR across all $\sigma$'s.

\textbf{Evaluation.}\quad We mainly evaluate model performance on two metrics: clean accuracy and average certified radius (ACR). Clean accuracy is the classification accuracy when taking the original test images as the input and cannot evaluate model robustness. A more reasonable metric for evaluating robustness is ACR. We follow the standard evaluation protocol used in~\cite{cohen2019certified,salman2019provably,zhai2019macer} for fair comparison: for each test data $(x_i, y_i)\in\mathcal{S}_{\text{test}}$, record the radius $R_i$ that can be certified by the model $g$. Set $R_i=0$ if $x_i$ can't be classified correctly by $g$. Then $\text{ACR}=\frac1{|\mathcal{S}_{\text{test}}|}\sum_{i}R_i$. Since the denominator is the size of the full test set, one cannot obtain large ACR without high accuracy. Thus ACR becomes a popular choice in most of the DL robustness literature. We use CERTIFY algorithm in \cite{cohen2019certified} to obtain certified radius and choose $N_0=100, N=100,000, \alpha=0.001$ in CERTIFY. 


\textbf{Baseline models.}\quad Four baseline models are discussed in this section: MACER~\cite{zhai2019macer}, SmoothAdv~\cite{salman2019provably}, SmoothMix~\cite{jeong2021smoothmix} and Constistency~\cite{jeong2020consistency}. For MACER, we follow the configurations given by Table 4 in the original paper \cite{zhai2019macer}. For SmoothAdv, we pick the best models under different $\sigma=0.25, 0.50, 1.00$ from the Github repository of \cite{salman2019provably}. See \cref{table_model_config} for more details on hyper-parameter selection of SmoothAdv. For SmoothMix and Consistency, we follow the same setting as in the original papers.
\input{tables/table_model_config}

\textbf{\ourmethod.}\quad We apply \cref{alg:macerADV} to train our \ourmethod models. On Cifar-10, we choose $\gamma=8.0$, $\lambda=12.0$, $\beta=16.0$ for all $\sigma=0.25, 0.50, 1.00$. The choices of $T, m, \epsilon$ are the same as SmoothAdv to ensure fair comparison and are summarized in \cref{table_model_config}. We follow the same training scheme as \cite{salman2019provably}. The initial learning rate is 0.1 and decays by a factor of 0.1 every 50 epochs. A batch size of 256 is used in the training. For more details, please refer to \cite{salman2019provably}. Note that by the choice of hyper-parameters, SmoothAdv and \ourmethod have the same training time, which implies the improved performance of \ourmethod is not gained from more expensive computation. The experiment results on Cifar-10 are summarized in \cref{table_main}.

Besides, the training time of three major models (SmoothAdv, MACER, \ourmethod) is reported in \cref{table_model_config}.  
Compared with SmoothAdv, \ourmethod has the same training cost but achieves both higher accuracy and higher ACR in all $\sigma$ (as in \cref{table_main}). Although \ourmethod has more training time for each epoch than MACER does, \ourmethod is still more computationally efficient due to fewer required training epochs.
For example, with $\sigma = 1$, MACER requires 440 epochs to obtain ACR 0.518 in 32.8 hrs, while \ourmethod only needs 150 epochs to achieve higher ACR 0.554 in 8 hrs. Thus, \ourmethod improves 7\% ACR while being 4$\times$ faster to train.
Training efficiency of \ourmethod compared with MACER can be observed in \cref{fig:acr-plot}.



\textbf{\EsbRs{}.}\quad We also employ our proposed ensemble techniques introduced in \cref{subsec:ensemble} to enhance robustness performance. \textcolor{blue}{\textbf{(I). }}For mixed-model ensemble, we use the following naming convention to report our result: \EsbRs{-}Model1$\times$n+Model2$\times$m represents the ensemble model obtained by $n$ Model1 and $m$ Model2. 
For example, \EsbRs{-}\ourmethod$\times$1+SmoothAdv$\times$2 represents the ensemble of one \ourmethod model and two SmoothAdv models.
Our empirical experiments in \cref{fig:ensemble} also verifies the theoretical analysis on the success of mixed-model ensemble in \cref{subsec:ensemble}. In ensemble experiments, we independently train all \ourmethod and SmoothAdv models on Cifar-10 with $\sigma=0.50$ and the model configuration is given by \cref{table_model_config}. In \cref{fig:ensemble}, 
we observe that mixed-model ensemble gives universally better ACR than intra-model ensemble (both perform better than non-ensemble baseline), which is in accordance to the analysis in \cref{subsec:ensemble}.
\textcolor{blue}{\textbf{(II). }}It can be seen from \cref{table_wt_ensemble} that all ensemble strategies increase ACR. Max margin ensemble (MME)~\cite{yang2021certified} slightly underperforms average weighted \EsbRs{}, while our optimal weighted \EsbRs{} can increase ACR by up to 4\%.

\input{tables/table_wt_ensemble}

\textbf{Experiment on ImageNet, SVHN and CIFAR}
We report accuracy and ACR of SmoothAdv, MACER and \ourmethod on ImageNet with $\sigma=0.25/0.50/1.00$. We train MACER model using the configuration provided by Table 5 of \cite{zhai2019macer}. For SmoothAdv, we choose the best configuration from \cite{salman2019provably}, that is $T=1,m=1,\epsilon=0.5$ and 90 epochs. For \ourmethod, we use the same configuration as SmoothAdv and set $\lambda=2,\gamma=8,\beta=16$ for these additional parameters. The performance and training cost is reported in \cref{table_imagenet}, from which we see that \ourmethod achieves best ACR in all $\sigma$ while being the most cost-efficient. The additional experiments results on SVHN are reported in \cref{app:svhn}. The certified accuracy table for Cifar-10 under $\sigma=0.25/0.50/1.00$ can be found in \cref{app_certified}.

\input{tables/table_imagenet}


\textbf{Performance and Discussion.}
\textcolor{blue}{\textbf{(I) Better performance.}} From \cref{table_main}, \cref{table_model_config}, \cref{table_imagenet} and \cref{table_svhn} in \cref{app:svhn}, we can conclude that our \ourmethod gives the best ACR among all $\sigma$'s and all SOTA baselines (SmoothAdv, MACER, SmoothMix, Consistency) on various datasets (Cifar-10, ImageNet, SVHN).
\textcolor{blue}{\textbf{(II) Higher efficiency.}} 
In terms of training cost, \ourmethod is 4x faster than MACER (while still achieving higher ACR); as fast as the most efficient SmoothAdv models but achieves higher ACR; 2.5x faster than Consistency (while still achieving larger ACR). 
\textcolor{blue}{\textbf{(III) Mixed-model ensemble. }}For ensemble models, \ourmethod$\times1$+SmoothAdv$\times2$ outperforms all the other models on Cifar-10 with $\sigma=0.50,1.00$, suggesting that one may prefer mixed-model ensemble in particular situations. Different from \cite{horvath2021boosting}, we are the first to observe mixed-model ensemble can outperform intra-model ensemble and perform analysis to explain this opposite phenomenon. Besides, the introduction of \ourmethod brings enriched diversity of component models, greatly improving the power of mixed-model ensemble and making \ourmethod and \EsbRs{} \textit{\textbf{integrated}} contributions.

%% file: tables/table_main_result_cifar.tex
\begin{table*}[t]
\caption{Cifar-10: ACR on 500 test images of Cifar-10. Clean accuracy is reported in parenthesis.}
\label{table_main}
\vskip 0.15in
\begin{center}
\begin{small}
\scalebox{0.8}{
\begin{tabular}{clcccc}
\toprule
& Methods & $\sigma=0.25$ & $\sigma=0.5$ & $\sigma=1.0$ & Ensemble?\\
\midrule
\multirow{4}{4em}{\centering Baselines}
&SmoothAdv~\cite{salman2019provably} & 0.541 (74.2\%) & 0.735 (56.4\%) & 0.758 (45.8\%) & $\times$\\
&MACER~\cite{zhai2019macer} &	0.518 (79.4\%) & 0.682 (63.4\%) & 0.768 (42.4\%)& $\times$\\
&SmoothMix~\cite{jeong2021smoothmix} &	0.545 (76.0\%) & 0.685 (63.8\%) & 0.626 (48.4\%) & $\times$\\
&Consistency~\cite{jeong2020consistency} & 0.535 (78.4\%) & 0.701 (64.6\%) & 0.719 (45.8\%) & $\times$\\
\midrule
\multirow{5}{4em}{\centering Ours}
&\ourmethod & \textbf{0.554} (76.0\%) & \textbf{0.742} (58.4\%) & \textbf{0.794} (47.6\%)& $\times$\\
&\EsbRs{-}\ourmethod$\times3$ & \textbf{0.583} (76.4\%) & 0.772 (58.8\%)& 0.805 (47.6\%) & $\surd$\\
&\EsbRs{-}\ourmethod$\times$1+SmoothAdv$\times$2 & 0.572 (77.2\%) & \textbf{0.783} (59.4\%) & \textbf{0.810} (47.2\%)& $\surd$\\
&\EsbRs{-}\ourmethod$\times$2+MACER$\times$1 & 0.568 (79.8\%) & 0.728 (63.6\%) & 0.801 (42.8\%)& $\surd$\\
&\EsbRs{-}\ourmethod$\times$1+MACER$\times$2 & 0.570 (80.4\%) & 0.723 (65.0\%) & 0.760 (44.0\%)& $\surd$\\
\bottomrule
\end{tabular}
}
\end{small}
\end{center}
\vskip -0.1in
\end{table*}

%% file: tables/table_model_config.tex
\begin{table*}[t]
\caption{Model configuration and training cost: main hyper-parameters and training time for SmoothAdv, MACER and \ourmethod on Cifar-10 with varing $\sigma$. For the additional parameters in \ourmethod, we pick $\lambda=12.0,\gamma=8.0,\beta=16.0$. ACR is also reported for easy comparison.}
\label{table_model_config}
\vskip 0.15in
\begin{center}
\begin{small}
\scalebox{0.85}{
\begin{tabular}{lcccccr}
\toprule
Models & $T$ & $m$ & $\epsilon$ & Epochs & ACR & Time\\
\midrule
SmoothAdv/\ourmethod ($\sigma=0.25$) & 2 & 8 & 1.0 & 150 & 0.541/\textbf{0.554} & 15.5h\\
SmoothAdv/\ourmethod ($\sigma=0.50$) & 2 & 8 & 2.0 & 150  & 0.735/\textbf{0.742} & 15.5h\\
SmoothAdv/\ourmethod ($\sigma=1.00$) & 2 & 4 & 2.0 & 150  & 0.758/\textbf{0.794} & 8h\\
\midrule
MACER ($\sigma=0.25/0.50/1.00$) & NA & 16 & NA & 440 & 0.518/0.682/0.768 & 32.8h\\
\bottomrule
\end{tabular}
}
\end{small}
\end{center}
\end{table*}

%% file: tables/table_wt_ensemble.tex
\begin{table}[t]
\caption{Optimal weighted 2-ensemble. ACR and clean accuracy on 500 test images of Cifar-10 with $\sigma=0.25$. All certification has parameters $N_0=100,N=100,000,\alpha=0.001$. Optimal weights are computed from \cref{alg:wt_ensemble} with \ourmethod models $F^1,F^2$, $m=10,n=10,t=0.3,\sigma=0.25,\tilde\sigma=0.01.$ }
\label{table_wt_ensemble}
\vskip 0.15in
\begin{center}
\begin{small}
\scalebox{0.9}{
\begin{tabular}{lccr}
\toprule
Model & Accuracy & ACR & Certificate Time\\
\midrule
\ourmethod (no ensemble) & 0.760 & 0.554 & 8.9s\\
Avg wt (baseline) & 0.760 & 0.572 & 18.0s\\
MME \cite{yang2021certified} (baseline) & 0.754 & 0.567 & 19.6s\\
\textbf{Optimal weight} (ours) & \textbf{0.766} & \textbf{0.576} & 26.3s\\
\bottomrule
\end{tabular}
}
\end{small}
\end{center}
\end{table}

%% file: tables/table_imagenet.tex

\begin{table}[th!]
\caption{ImageNet: ACR on 500 test images of ImageNet. Clean accuracy is reported in parenthesis.}
\label{table_imagenet}
\vskip 0.15in
\begin{center}
\begin{small}
\scalebox{0.75}{
\begin{tabular}{llcccc}
\toprule
Methods & $\sigma=0.25$ & $\sigma=0.5$ & $\sigma=1.0$ & Time\\
\midrule
SmoothAdv & 0.519 (61.5\%) & 0.801 (55.6\%) & 0.971 (41.4\%) & 48h\\
MACER &	0.438 (63.2\%) & 0.628 (52.6\%) & 0.634 (37.8\%) & 70h\\
\midrule
\ourmethod (ours) & \textbf{0.537 (63.9\%)} & \textbf{0.837 (56.2\%)} & \textbf{0.989 (45.6\%)} & 48h\\
\bottomrule
\end{tabular}
}
\end{small}
\end{center}
\vskip -0.1in
\end{table}

%% file: 5_conclusion.tex
\section{Conclusions}
We have proposed two novel and cost-effective approaches to promote robustness of randomized smoothed classifiers: \ourmethod improve the robustness by maximizing the certified radius over adversarial example, and \EsbRs{} can further improve \ourmethod on both clean accuracy and robustness certificate. Through extensive numerical experiments, we show that \ourmethod outperforms major baseline models (SmoothAdv, MACER, Consistency, SmoothMix) on various datasets (Cifar-10, ImageNet, SVHN).
The ACR improvement is up to 15\% compared with MACER and 8\% compared with the best models of SmoothAdv. Moreover, we provided a general theoretical analysis for \EsbRs{} and develop a theoretical-grounded methodology to design optimal ensemble scheme, which outperforms prior works.

\section*{Acknowledgement} T.-W. Weng is supported by National
Science Foundation under Grant No. 2107189.

%% file: appendix.tex

\newpage
\onecolumn

\begin{center}
    \textbf{{{Appendix}}}
\end{center}

\section{Full algorithm of \ourmethod}
\label{app:alg}

\input{algorithms/alg_macer-adv}

\newpage
\section{Full algorithm of optimal weight design}
\label{app:wt_ensemble}
\input{algorithms/alg_opt.tex}



\section{Certified accuracy}
\label{app_certified}
We also provide certified accuracy table for Cifar-10, which is presented in \cref{table_certified_acc_0.25}, \cref{table_certified_acc} and \cref{table_certified_acc_1.00}.
\input{tables/table_certified_acc_0.25}
\input{tables/table_certified_acc}
\input{tables/table_certified_acc_1.00}


\section{Additional experiments on SVHN dataset}
\label{app:svhn}
We compare the performance of SmoothAdv, MACER and \ourmethod on SVHN dataset with $\sigma=0.25, 0.50$.
On SVHN with $\sigma=0.25$, we choose $T=2$, $m=4$, $\lambda=12.0$, $\gamma=8.0$, $\beta=16.0$, $\epsilon=0.5$ and train the model for 150 epochs. On SVHN with $\sigma=0.50$, we still choose $T=2$, $m=4$, $\gamma=8.0$, $\beta=16.0$, $\epsilon=0.5$ but a different $\lambda=4.0$. The model is also trained for 150 epochs. The initial learning rate is set to 0.01 and drops by a factor of 0.1 every 50 epochs. The other training details follow the same as Cifar-10. For SmoothAdv, take $T=2, m=4, \epsilon=0.5$ when $\sigma=0.25$ and $T=2, m=4, \epsilon=0.25$ when $\sigma=0.50$. We train MACER model for 440 epochs whose configuration is given by C.2.2 of \cite{zhai2019macer}. We report the experiment results in \cref{table_svhn}.
\input{tables/table_svhn}

\section{Proof of \cref{thm_ours}}
\label{app_proof}
For completeness, recall that we have $k$ trained soft classifiers $F^1,\dots,F^k:\mb{R}^d\to P(\mathcal{Y})$ and $\mathcal{Y}=\{1,\dots,c\}$. Consider soft-ensemble model $H$ whose output is a weighted average of the probabilities from $F^1,\dots,F^k$:
$H(x)=\sum_{l=1}^kw_l(x)F^l(x).$ Suppose the associated hard classifier is
$h(x)=\argmax_{c\in\mathcal{Y}}\big(H(x)\big)_c.$ Then we apply RS to $h$ and get the corresponding smoothed classifier $e(x)=\argmax_c\mP_{\epsilon\sim\mathcal{N}(0,\sigma^2I)}(h(x+\epsilon)=c)$. It can be shown that the RS classifier $e$ of an ensemble model also has the same certification guarantee as in \cite{cohen2019certified}, which is presented in the following theorem.

\begin{theorem}[Robustness gaurantee for ensemble]
Suppose that under Gaussian perturbation $\epsilon\sim\mathcal{N}(0, \sigma^2I)$, the most likely class $c_A$ is returned by the 
\EsbRs{} $e$ with probability $p_A$ and the second most likely class $c_B$ is returned with probability $p_B$,
then we have $e(x+\delta)=e(x)$ for all $\|\delta\|_2\leq R$, where $R = \frac\sigma2(\Phi^{-1}({ p_A})-\Phi^{-1}({ p_B})).$
\end{theorem}
Without loss of generality, we assume $\sigma=1$. The proof follows similarly for $\sigma\neq1$. Before presenting the formal proof, we proceed with several preparatory lemmas from \cite{salman2019provably}.
\begin{theorem}[Lemma 2 from \cite{salman2019provably}]\label{lem_salman}
Let $f:\mb{R}^n\to[0, 1]$ be a deterministic (non-random) function and define $\hat f$ by 
$$\hat f(x) = (f * \mathcal{N}(0, I))(x) = \frac1{(2\pi)^{n/2}}\int_{\mb{R}^n}f(t)\exp\bigg(-\frac12\|x-t\|^2\bigg)\,dt.$$ 
Let $\Phi(a)=\frac1{\sqrt{2\pi}}\int_{-\infty}^a\exp\Big(-\frac12s^2\Big)\,ds.$
Then the map $x\mapsto\Phi^{-1}(\hat f(x))$ is 1-Lipschitz.
\end{theorem}
We refer the readers to \cite{salman2019provably} for the proof of \cref{lem_salman}. Now we are ready to prove \cref{thm_ours}.

\begin{proof} [Proof of \cref{thm_ours}]
    Let $f_i(x)=\mathbb{I}(e(x)=i)$ for $i\in\mathcal{Y}=\{1,2,\dots, c\}.$ Define 
    \begin{equation*}
\label{eq:fhat}
    \hat f_i(x)= (f_i * \mathcal{N}(0, I))(x) = \mP_{\epsilon\sim\mathcal{N}(0, I)}(e(x+\delta)=c_i).
\end{equation*}
From this definition, we immediately have $p_A=\hat f_A(x)$ and $p_B=\hat f_B(x)$.
Then by \cref{lem_salman}, we have $\Phi^{-1}(\hat f_i(x))$ is 1-Lipschitz for all $i\in\{1,\dots,c\}$. So for any $\delta$,
    \begin{equation*}
        \Phi^{-1}(\hat f_A(x)) - \Phi^{-1}(\hat f_A(x+\delta))\leq\|\delta\|
    \end{equation*}
    Suppose $\delta^*$ is a successful adversarial noise, namely $\hat f_A(x+\delta^*)\leq\hat f_B(x+\delta^*)$ for some class $c_B$, then
    \begin{equation}
    \label{eq:r1}
            \Phi^{-1}(\hat f_A(x)) - \Phi^{-1}(\hat f_B(x+\delta^*))\leq\Phi^{-1}(\hat f_A(x)) - \Phi^{-1}(\hat f_A(x+\delta^*))\leq\|\delta^*\|.
    \end{equation}
    Apply \cref{lem_salman} again to $\hat f_B$ and we obtain
    \begin{equation}
    \label{eq:r2}
            \Phi^{-1}(\hat f_B(x+\delta^*)) - \Phi^{-1}(\hat f_B(x))\leq\|\delta^*\|.
    \end{equation}
    Combining \cref{eq:r1} and \cref{eq:r2}, we have that if $\delta^*$ is a successful attack, it has to satisfy
    $$\|\delta^*\|\geq\frac12(\Phi^{-1}(p_A)-\Phi^{-1}(p_B)).$$
    In other words, we are guaranteed that $e(x)=e(x+\delta)$ if 
    $$\|\delta\|_2\leq\frac12(\Phi^{-1}(p_A)-\Phi^{-1}(p_B)).$$
\end{proof}

%% file: algorithms/alg_macer-adv.tex
\begin{algorithm}[h!]
\caption{Our \textbf{\ourmethod}$(\sigma,m,T,\lambda,\beta,\gamma)$}
\label{alg:macerADV}
\begin{algorithmic}
   \State {\bfseries Input:} training set $\hat p_{\text{data}}$, noise level $\sigma$, number of Gaussian samples $m$, regularization parameter $\lambda$, hinge factor $\gamma$, inverse temperature $\beta$, number of PGD step $T$ 
   \For{each iteration}
    \State 1) Sample a mini-batch $(x_1, y_1),\dots,(x_n,y_n)\sim\hat p_{\text{data}}$
    \State 2) For each $(x_i, y_i)$, use $T$-step SmoothAdv to generate adversarial example $\hat x_i$
    \State 3) For each $(\hat x_i, y_i)$, draw $m$ i.i.d. Gaussian samples $x_{i1},\dots,x_{im}$ from $\mathcal{N}(x_i,\sigma^2I)$
    \State 4) Obtain an estimation of $G_\theta(\hat x)$ by $\hat z_\theta(\hat x) \gets \frac1m\sum_{k=1}^mF_\theta(\hat x_{ik}), \text{ for } i=1,\dots, n$
    \State 5) Collect the set of data with correct prediction: $\mathcal{S}_\theta=\{i: y_i=\argmax_c\hat z_\theta(\hat x_i)_c\}$
    \State 6) For each $i\in\mathcal{S}_\theta$, compute the second most likely class $\hat y_i \gets \argmax_{c\neq y_i}\hat z_\theta(\hat x_i)_c$
    \State 7) For each $i\in\mathcal{S}_\theta$, compute $\hat\xi(\hat x_i, y_i) \gets \Phi^{-1}(\hat z_\theta(x)_{y_i}) - \Phi^{-1}(\hat z_\theta(x)_{\hat y_i})$
    \State 8) Sample $\delta\sim\mathcal{N}(0,\sigma^2I)$ and update $\theta$ with SGD to minimize
    \begin{align*}
        -\frac1n\sum_{i=1}^n\log\hat z_\theta(\hat x_i+\delta)_{y_i}
        +\frac{\lambda\sigma}{2n}\sum_{i\in\mathcal{S}_\theta}\max\{\gamma-\hat\xi_\theta(\hat x_i+\delta,y_i), 0\}
    \end{align*}
    \EndFor   
    \State {\bfseries Output:} model parameters $\theta$
\end{algorithmic}
\end{algorithm}

%% file: algorithms/alg_opt.tex
\algdef{SE}[SUBALG]{Indent}{EndIndent}{}{\algorithmicend\ }%
\algtext*{Indent}
\algtext*{EndIndent}

\begin{algorithm*}[th!]
\caption{Optimal weights of ensemble with 2 models. The function \textbf{ComputeWeight} will return the optimal weights}
\label{alg:wt_ensemble}
\begin{algorithmic}
   \State {\bfseries Input:} two base models $F^1,F^2$, number of Gaussian noise $n$, number of perturbed models $m$, noise level $\sigma$, proportion of the parameters to perturb $t$, standard deviation of perturbation on parameters $\tilde\sigma$, query point $x$, target label $y$
   \State{\bfseries function} PerturbModel($F^1,F^2,m,t,\tilde\sigma$):
   \Indent
    \For {$l=1,2$}
       \For{each $j=1,\dots,m$}
        \For {each parameter $\theta$ in $F^l$}
            \State Draw a Bernoulli variable $X$ from $\text{Bernoulli}(t)$
            \If{$X=1$} 
            \State Draw $\delta\sim\mathcal{N}(0,\tilde\sigma^2)$ and update $\theta\gets\theta+\delta$
            \EndIf
        \EndFor
        \State Store perturbed model $\hat F^{l}_j$
        \EndFor
        \EndFor
        \State {\bfseries Output:} perturbed models $\hat F^1_1,\dots,F^1_m$ and $\hat F^2_1,\dots,\hat F^2_m$.
    \EndIndent
    \State
    \State{\bfseries function} Estimation([$\hat F^1_1,\dots,F^1_m$], [$\hat F^2_1,\dots,\hat F^2_m$], $\sigma,n,x,y$):
    \Indent
        \State Draw $n$ i.i.d. noisy samples from $\mathcal{N}(x,\sigma^2I)$ and denote them by $x_1,\dots,x_n$
        \For{$l=1,2$}
        \For{$i=1,\dots,m$}
            \For{$j=1,\dots,n$}
                \State Compute $z^{l,(i-1)n+j}\gets\hat F^l_i(x_j)_y\mathbf{1}-\hat F^l_i(x_j)$, where $\mathbf{1}=[1,1,\dots,1]^\top$ is the vector of all 1's.
            \EndFor
        \EndFor
        \EndFor
        \State{\bfseries Output:} estimates of logits $z^{l,j}$ for $l=1,2$ and $j=1,\dots,mn$
    \EndIndent
    \State
    \State{\bfseries function} ComputeWeight($F^1,F^2,\sigma,\tilde\sigma,n,m,t,x,y$):
    \Indent
        \State 1) $[\hat F^1_1,\dots,F^1_m]$, $[\hat F^2_1,\dots,\hat F^2_m]$ $\gets$ PerturbModel($F^1,F^2,m,t,\tilde\sigma$)
        \State 2) $z^{l,j}\gets$ Estimation([$\hat F^1_1,\dots,F^1_m$], [$\hat F^2_1,\dots,\hat F^2_m$], $\sigma,n,x,y$) for $l=1,2$ and $j=1,\dots,m$
        \State 3) Compute $\bar z^l\gets\frac1{mn}\sum_{j=1}^{mn}z^{l,j}$ for $l=1,2$ and $a_i\gets\min\{\bar z^1_i,\bar z^2_i\}^{-2}$ for $i=1,\dots,c$
        \State 4) Compute 
                 $$   b_i\gets\frac1{mn}\sum_{j=1}^{mn}(z^{1,j}-\bar z^{1})(z^{1,j}-\bar z^{1})^\top_{ii},~
                    c_i\gets\frac2{mn}\sum_{j=1}^{mn}(z^{1,j}-\bar z^{1})(z^{2,j}-\bar z^{2})^\top_{ii},~
                    d_i\gets\frac1{mn}\sum_{j=1}^{mn}(z^{2,j}-\bar z^{2})(z^{2,j}-\bar z^{2})^\top_{ii},$$
        \State 5) Compute $$ A=\sum_{i=2}^ca_i(b_i+c_i+d_i),\quad
                B=\sum_{i=2}^c -a_i(c_i+2d_i),\quad
                C=\sum_{i=2}^ca_id_i.$$
        \If{$A>0$ and $0\leq-\frac{B}{2A}\leq1$}
        \State $w_1\gets-\frac{B}{2A}$ and $w_2\gets1+\frac{B}{2A}$
        \Else
        \State $w_1\gets0,w_2\gets1$ if $A+B>0$ else $w_1\gets1,w_2\gets0$
        \EndIf
        \State{\bfseries Output:} $w_1,w_2$
    \EndIndent
\end{algorithmic}
\end{algorithm*}

%% file: tables/table_certified_acc_0.25.tex
\begin{table*}[h]
\caption{Certified accuracy: certified accuracy and ACR of the first 500 test images of Cifar-10 with $\sigma=0.25$. Each column represents the robust accuracy that can be certified at this $\ell_2$ radius.}
\label{table_certified_acc_0.25}
\vskip 0.15in
\begin{center}
\begin{small}
\scalebox{1}{
\begin{tabular}{lcccccccccr}
\toprule
Model ($\sigma=0.25$) & 0.00 & 0.25 & 0.50 & 0.75 & 1.00 & 1.25 & 1.50 & 1.75 & 2.00 & ACR\\
\midrule
SmoothAdv & 0.742 & 0.660 & \textbf{0.572} & 0.45 & 0.0 & 0.0 & 0.0 & 0.0 & 0.0  & 0.541\\
MACER & \textbf{0.794} & \textbf{0.678} & 0.524 & 0.400 & 0.0 & 0.0 & 0.0 & 0.0 & 0.0 &  0.518\\
SmoothMix & 0.76 & 0.688 & 0.572 & 0.446 & 0.0 & 0.0 & 0.0 & 0.0 & 0.0 & 0.545\\
\ourmethod & 0.76 & 0.668 & \textbf{0.572} & \textbf{0.484} & 0.0 & 0.0 & 0.0 & 0.0 & 0.0 &  \textbf{0.554}\\
\midrule
\EsbRs{-}\ourmethod$\times3$ & 0.764 & 0.700 & \textbf{0.614} & \textbf{0.514} & 0.0 & 0.0 & 0.0 & 0.0 & 0.0 &  \textbf{0.583}\\
\EsbRs{-}SmoothAdv$\times3$ & 0.766 & 0.698 & 0.600 & 0.506 & 0.0 & 0.0 & 0.0 & 0.0 & 0.0 &  0.576\\
\EsbRs{-}\ourmethod$\times$1+SmoothAdv$\times$2 & 0.772 & 0.672 & 0.594 & 0.498 & 0.0 & 0.0 & 0.0 & 0.0 & 0.0 &  0.572\\
\EsbRs{-}\ourmethod$\times$2+MACER$\times$1 & 0.798 & 0.700 & 0.586 & 0.472 & 0.0 & 0.0 & 0.0 & 0.0 & 0.0 &  0.568\\
\EsbRs{-}\ourmethod$\times$1+MACER$\times$2 & \textbf{0.804} & \textbf{0.714} & 0.598 & 0.462 & 0.0 & 0.0 & 0.0 & 0.0 & 0.0 &  0.570\\
\bottomrule
\end{tabular}
}
\end{small}
\end{center}
\end{table*}

%% file: tables/table_certified_acc.tex
\begin{table*}[h]
\caption{Certified accuracy: certified accuracy and ACR of the first 500 test images of Cifar-10 with $\sigma=0.50$. Each column represents the robust accuracy that can be certified at this $\ell_2$ radius.}
\label{table_certified_acc}
\vskip 0.15in
\begin{center}
\begin{small}
\scalebox{1}{
\begin{tabular}{lcccccccccr}
\toprule
Model ($\sigma=0.50$) & 0.00 & 0.25 & 0.50 & 0.75 & 1.00 & 1.25 & 1.50 & 1.75 & 2.00 & ACR\\
\midrule
SmoothAdv & 0.564 & 0.516 & 0.468 & 0.432 & 0.394 & 0.328 & \textbf{0.286} & \textbf{0.224} & 0.0 & 0.735\\
MACER & \textbf{0.634} & \textbf{0.566} & 0.476 & 0.432 & 0.346 & 0.258 & 0.206 & 0.126 & 0.0 & 0.682\\
SmoothMix & 0.638 & 0.548 & 0.48 & 0.416 & 0.34 & 0.274 & 0.21 & 0.152 & 0.0 & 0.685\\
\ourmethod & 0.584 & 0.532 & \textbf{0.486} & \textbf{0.442} & \textbf{0.398} & \textbf{0.334} & 0.270 & 0.216 & 0.0 & \textbf{0.742}\\
\midrule
\EsbRs{-}\ourmethod$\times3$ & 0.588 & 0.544 & 0.498 & 0.448 & 0.414 & 0.360 & 0.288 & 0.230 & 0.0 & 0.772\\
\EsbRs{-}SmoothAdv$\times3$ & 0.584 & 0.530 & 0.476 & \textbf{0.454} & 0.420 & 0.362 & 0.308 & \textbf{0.254} & 0.0 & 0.777\\
\EsbRs{-}\ourmethod$\times$1+SmoothAdv$\times$2 & 0.594 & 0.540 & 0.482 & \textbf{0.454} & \textbf{0.422} & \textbf{0.374} & \textbf{0.310} & 0.238 & 0.0 & \textbf{0.783}\\
\EsbRs{-}\ourmethod$\times$2+MACER$\times$1 & 0.636 & 0.564 & 0.504 & 0.446 & 0.388 & 0.294 & 0.240 & 0.172 & 0.0 & 0.728\\
\EsbRs{-}\ourmethod$\times$1+MACER$\times$2 & \textbf{0.650} & \textbf{0.568} & \textbf{0.506} & 0.450 & 0.370 & 0.292 & 0.220 & 0.158 & 0.0 & 0.723\\
\bottomrule
\end{tabular}
}
\end{small}
\end{center}
\end{table*}

%% file: tables/table_certified_acc_1.00.tex
\begin{table*}[h]
\caption{Certified accuracy: certified accuracy and ACR of the first 500 test images of Cifar-10 with $\sigma=1.00$. Each column represents the robust accuracy that can be certified at this $\ell_2$ radius.}
\label{table_certified_acc_1.00}
\vskip 0.15in
\begin{center}
\begin{small}
\scalebox{1}{
\begin{tabular}{lccccccccccr}
\toprule
Model ($\sigma=1.00$) & 0.00 & 0.25 & 0.50 & 0.75 & 1.00 & 1.25 & 1.50 & 1.75 & 2.00 & 2.25 & ACR\\
\midrule
SmoothAdv & 0.458 & 0.418 & 0.374 & 0.312 & 0.288 & 0.254 & 0.234 & 0.196 & 0.180 & 0.158 & 0.758\\
MACER & 0.424 & 0.392 & 0.354 & 0.328 & \textbf{0.304} & \textbf{0.274} & \textbf{0.250} & \textbf{0.212} & 0.184 & 0.156 & 0.768\\
SmoothMix & 0.484 & 0.42 & 0.348 & 0.292 & 0.244 & 0.22 & 0.178 & 0.15 & 0.124 & 0.096 & 0.626\\
\ourmethod & \textbf{0.476} & \textbf{0.440} & \textbf{0.392} & \textbf{0.350} & 0.302 &\textbf{0.274} & 0.236 & \textbf{0.212} & \textbf{0.186} & \textbf{0.164} & \textbf{0.794}\\
\midrule
\EsbRs{-}\ourmethod$\times3$ & \textbf{0.476} & 0.426 & 0.386 & 0.358 & 0.302 & 0.270 & 0.246 & 0.220 & 0.196 & 0.172 & 0.805\\
\EsbRs{-}SmoothAdv$\times3$ & 0.466 & \textbf{0.432} & 0.388 & \textbf{0.360} & 0.294 & 0.260 & 0.240 & 0.212 & 0.186 & 0.170 & 0.801\\
\EsbRs{-}\ourmethod$\times$1+SmoothAdv$\times$2 & 0.472 & \textbf{0.432} & \textbf{0.394} & 0.356 & 0.304 & 0.262 & 0.240 & 0.214 & 0.194 & \textbf{0.174} & \textbf{0.810}\\
\EsbRs{-}\ourmethod$\times$2+MACER$\times$1 & 0.428 & 0.404 & 0.370 & 0.342 & \textbf{0.318} & 0.276 & \textbf{0.256} & \textbf{0.230} & \textbf{0.198} & 0.164 & 0.801\\
\EsbRs{-}\ourmethod$\times$1+MACER$\times$2 & 0.440 & 0.404 & 0.366 & 0.340 & 0.304 & \textbf{0.282} & 0.238 & 0.202 & 0.184 & 0.154 & 0.760\\
\bottomrule
\end{tabular}
}
\end{small}
\end{center}
\end{table*}

%% file: tables/table_svhn.tex
\begin{table*}[h!]
\caption{SVHN: clean accuracy and ACR of different models evaluated on 500 test images of SVHN with varing $\sigma$.}
\label{table_svhn}
\vskip 0.15in
\begin{center}
\begin{small}

\begin{tabular}{lccccr}
\toprule
 $\sigma$ & Model & Accuracy & ACR & Training Time\\
\midrule
\multirow{9}{2em}{$0.25$}
& SmoothAdv & 85.8\% & 0.560 & 11.4h\\
& MACER & 86.8\% & 0.549 & 48.5h\\
& \ourmethod & 86.6\% & \textbf{0.569} & 11.4h\\
\cmidrule{2-5}
& \EsbRs{-}SmoothAdv$\times$3 & 87.8\% & 0.578 & NA\\
& \EsbRs{-}\ourmethod$\times$3 & 88.2\% & \textbf{0.582} & NA\\
& \EsbRs{-}\ourmethod$\times$1+MACER$\times$2 & 87.8\% & 0.559 & NA\\
& \EsbRs{-}\ourmethod$\times$2+MACER$\times$1 & 88.6\% & 0.570 & NA\\
& \EsbRs{-}\ourmethod$\times1$+SmoothAdv$\times2$ & 87.8\% & 0.577 & NA\\
& \EsbRs{-}\ourmethod$\times2$+SmoothAdv$\times1$ & 87.6\% & \textbf{0.582} & NA\\
\midrule
\multirow{9}{2em}{$0.50$}
& SmoothAdv & 71.2\% & 0.552 & 11.4h\\
& MACER & 58.4\% & 0.535 & 48.5h\\
& \ourmethod & 67.8\% & \textbf{0.572} & 11.4h\\
\cmidrule{2-5}
& \EsbRs{-}SmoothAdv$\times$3 & 71.2\% & 0.573 & NA\\
& \EsbRs{-}\ourmethod$\times$3 & 70.4\% & \textbf{0.588} & NA\\
& \EsbRs{-}\ourmethod$\times$1+MACER$\times$2 & 62.8\% & 0.551 & NA\\
& \EsbRs{-}\ourmethod$\times$2+MACER$\times$1 & 66.0\% & 0.564 & NA\\
& \EsbRs{-}\ourmethod$\times1$+SmoothAdv$\times2$ & 71.8\% & 0.577 & NA\\
& \EsbRs{-}\ourmethod$\times2$+SmoothAdv$\times1$ & 71.2\% & 0.583 & NA\\
\bottomrule
\end{tabular}

\end{small}
\end{center}
\end{table*}